\documentclass[conference]{IEEEtran}
\IEEEoverridecommandlockouts

\usepackage[utf8]{inputenc}
\usepackage[T1]{fontenc}
\usepackage{epsfig}
\usepackage{geometry}
\usepackage{csquotes}
\usepackage[english]{babel}

\makeatletter
\DeclareRobustCommand{\cev}[1]{%
  \mathpalette\do@cev{#1}%
}
\newcommand{\do@cev}[2]{%
  \fix@cev{#1}{+}%
  \reflectbox{$\m@th#1\vec{\reflectbox{$\fix@cev{#1}{-}\m@th#1#2\fix@cev{#1}{+}$}}$}%
  \fix@cev{#1}{-}%
}
\newcommand{\fix@cev}[2]{%
  \ifx#1\displaystyle
    \mkern#23mu
  \else
    \ifx#1\textstyle
      \mkern#23mu
    \else
      \ifx#1\scriptstyle
        \mkern#22mu
      \else
        \mkern#22mu
      \fi
    \fi
  \fi
}

\makeatother

\usepackage{mathtools}

\title{\LARGE \bf An In-depth Walkthrough on Evolution of Neural Machine Translation}

\begin{document}
\author{
    \IEEEauthorblockN{Rohan Jagtap\textsuperscript{1}, Dr. Sudhir N. Dhage \textsuperscript{2}}
    \IEEEauthorblockA{Department of Computer Engineering, Sardar Patel Institute of Technology\\Mumbai, India\\}
    \IEEEauthorblockA{rohan.jagtap@spit.ac.in\textsuperscript{1}, sudhir\_dhage@spit.ac.in\textsuperscript{2}}
}
\maketitle
\thispagestyle{empty}
\pagestyle{empty}

\begin{abstract}
Neural Machine Translation (NMT) methodologies have burgeoned from using simple feed-forward architectures to the state of the art; viz. BERT model. The use cases of NMT models have been broadened from just language translations to conversational agents (chatbots), abstractive text summarization, image captioning, etc. which have proved to be a gem in their respective applications.  This paper aims to study the major trends in Neural Machine Translation, the state of the art models in the domain and a high level comparison between them.
\end{abstract}



\section{INTRODUCTION}

Machine Translation is the branch of computational linguistics which focuses on concepts dealing with translation of text or speech from one language to another (nowadays from one 'form' to another). These concepts have proved to be a very significant leap in computational linguistics and have a rich set of use cases including cross platform applications, international communication, medical domain, etc. The origins of machine translations come from substituting the tantamount words from one language to other. This was followed by rule based approaches which include eliciting several protocols based on the respective grammatical senses for the translation process. These methods just quite serve the purpose but substitution of words may change the meaning of the context or equivocal words may add to the confusion. With the rise of Machine Learning paradigms and statistical modelling, Statistical Machine Translation was born. These concepts use information theory to predict a word in the target language from existing bilingual corpora with the help of probabilities. However, the collection of these corpora may be expensive. Also these methods have been proved to fail to an extent while translating between languages that do not follow similar grammatical patterns (e.g. ordering of verbs / adjectives / nouns).

The state of the art Language Modelling concepts were engendered and disseminated with Neural Networks. In the age of Deep Learning, Artificial Neural Networks have known to outperform almost all the statistical models in Machine Translation Domain. Neural Machine Translation is hence Machine Translation using Artificial Neural Networks. In this paper, we discuss the major leaps in NMT, various architectures, evaluation metrics, benefits, shortcomings and comparison among the models. 

\section{Recurrent Neural Networks}
In Regular Feed Forward Neural Networks, the inputs are multiplied with respective weights to obtain intermediate hidden representations (layers) and finally the output. These hidden layers basically imply a very complex mathematical non-linear function. Hence a vanilla Feed Forward Neural Network maps the inputs to the outputs by establishing a mathematical relation between them. However here, the output is the function of just the current inputs. On the contrary, in Recurrent Neural Networks (RNNs), the hidden layers along with the current inputs, also consider the previous hidden layer outputs. These previous outputs introduce a notion of 'memory' in the Neural Network architecture. The use case of this type of architecture is when the inputs are in a form of a series where the inputs as well as the previous 'timestep' contribute to the output. A typical use case of this can be finding trends in a 'Time Series'. The backpropagation in these neural nets is called as 'Backpropagation Through Time' (BPTT). However, with BPTT, a major issue of Vanishing Gradients\cite{vanishing_grad} arose. These issues were overcome by Long Short-Term Memory Networks and its variants. So these recurring properties of RNNs can be exploited for Natural Language Processing tasks since in any Language Model, the words in a given context definitely depend on the previous one and the similarly for sentences, the pattern follows. This is the motivation behind utilizing the architectural benefits of RNNs in Neural Machine Translation.

\begin{figure}[!h]
\begin{center}
\scalebox{0.35}{\includegraphics{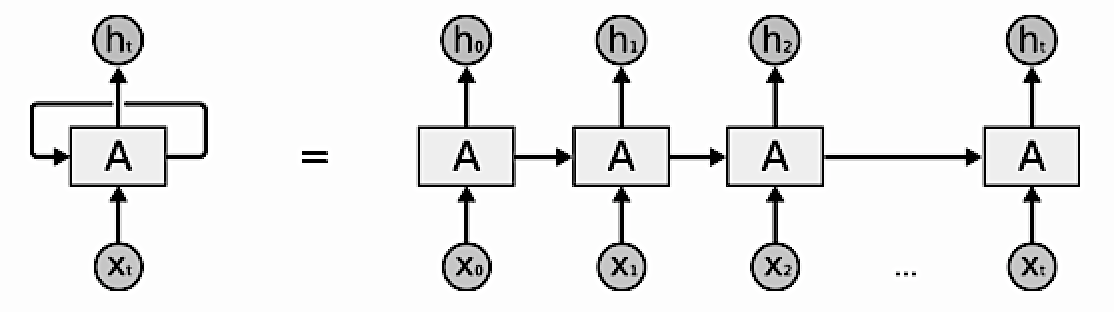}}
\caption {A typical Recurrent Neural Network Cell (with unrolled timesteps)}
\vspace{0mm}
\end{center}
\label{rnn}
\end{figure}

\section{The Sequence to Sequence Architecture}
Vanilla RNNs can be used in NMT, but it would be similar to a word-to-word mapping model, just using neural networks. The intuition to this can be obtained from the example of English to Chinese translation; the length of the English sequence may not be equal to the corresponding Chinese sequence. Also, the translation of an English word at some position in the sentence may not be at the same position in the Chinese context. Hence for such translations \cite{seq2seq} proposes the Sequence to Sequence architecture. It is an encoder-decoder architecture, i.e. the input sequence (English sentence in the above example) will be encoded first into an intermediate format (vector) and then using the encoded vector the target sequence (Chinese sentence) will be decoded. In the architecture proposed in \cite{seq2seq}, there are two RNNs (LSTMs), one as encoder and other as decoder respectively. The depth of the network maybe varied by increasing the number of LSTMs at each side. The cell states of LSTMs are are updated taking one word per timestep (depending on the use case), the output sequence of the encoder LSTM is superfluous, the cell states are passed on to the decoder and using those states the decoder predicts the target sequence one word per timestep. The intuition to this model can be obtained by reconciling it with a human translator; for translating a sentence, rather than translating each word of the sentence individually, one would first 'listen' to the complete sentence (sequence) and then translate it. 

\begin{figure}[!h]
\begin{center}
\scalebox{0.1}{\includegraphics{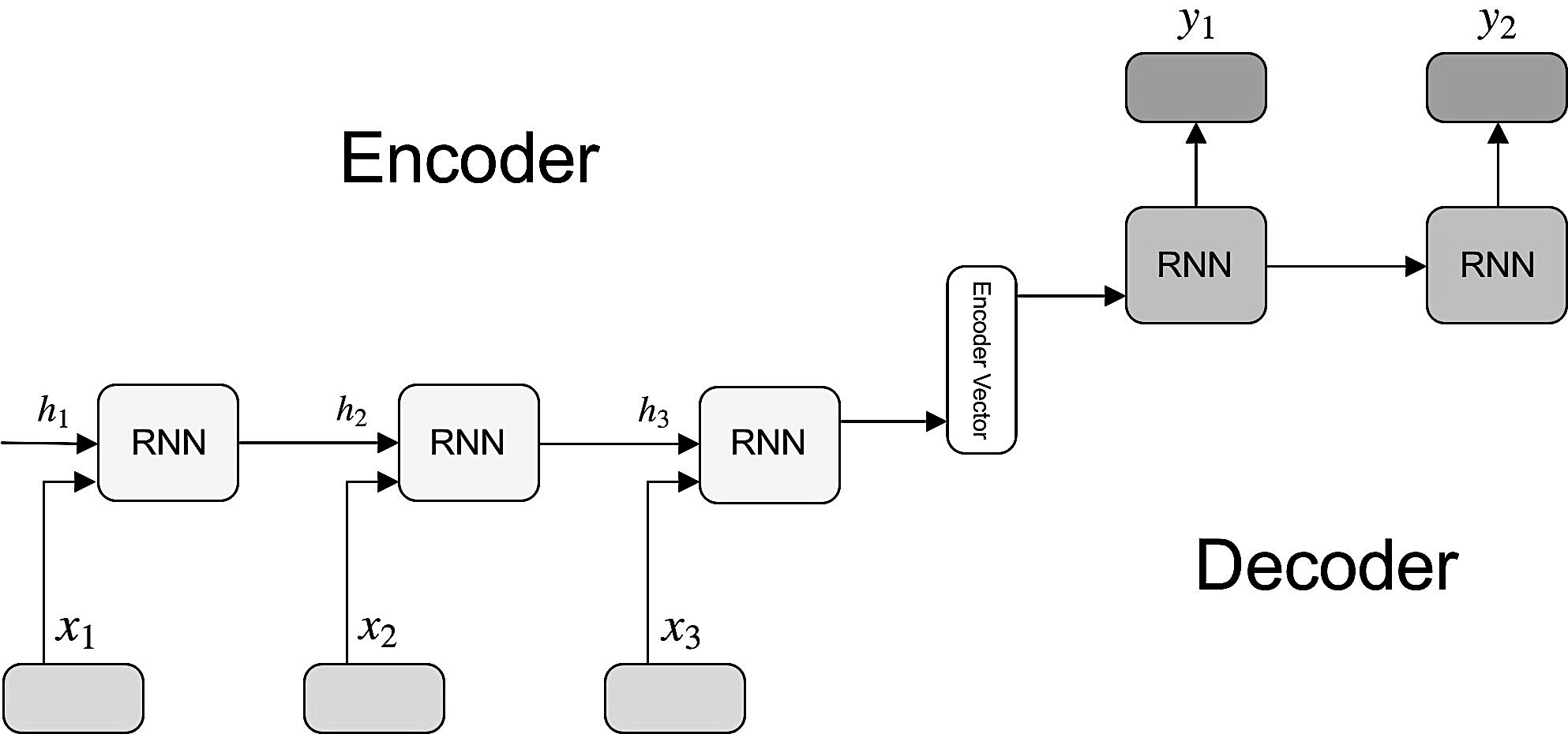}}
\caption {Sequence to Sequence Architecture \cite{seq2seq}}
\vspace{0mm}
\end{center}
\label{seq2seq}
\end{figure}

The Sequence to Sequence architecture is the base architecture for the state of the art NMT models. In \cite{seq2seq} the main dataset used is WMT’14 English to French; the BLEU\cite{bleu} score obtained was 34.81 by directly extracting translations from an ensemble of 5 deep LSTMs (with 384M parameters and 8,000 dimensional state each) using a simple left-to-right beam-search decoder. This was the best result for direct translations using Neural Network at the time.

\section{The Attention Mechanism}
The Attention Mechanism is an addition on the Sequence to Sequence Architecture\cite{seq2seq}. The most common implementation of attention is Bahadanau Attention\cite{bahadanau}.
The problem with sequence to sequence is that it is incapable of remembering long sequences. The intuition to this can be obtained by relating this to the fact that just a single cell state vector is being passed on through various timesteps of a LSTM cell and is updated at every timestep. In such a vector which is overwritten multiple times (in case of longer sequences), the data from the initial cell states may start to disappear. The name attention too, comes from the human tendency to pay 'attention' only on certain portions of a context while making translations from one context to another. Hence this suggests that rather than focusing on "How much to remember?", attention answers "What to remember?". The idea behind implementation of attention is that, all the intermediate context vectors should contribute to the final context vectors rather than just the last vector. 
\subsection{Bahadanau Attention}
\cite{bahadanau} suggests taking a weighted sum of all the intermediate context vectors from all the timesteps, the weights being an 'alignment score' of those vectors; the mechanism is mathematically represented in equations (1), (2), (3) and (4).
\begin{figure}[!h]
\begin{center}
\scalebox{0.6}{\includegraphics{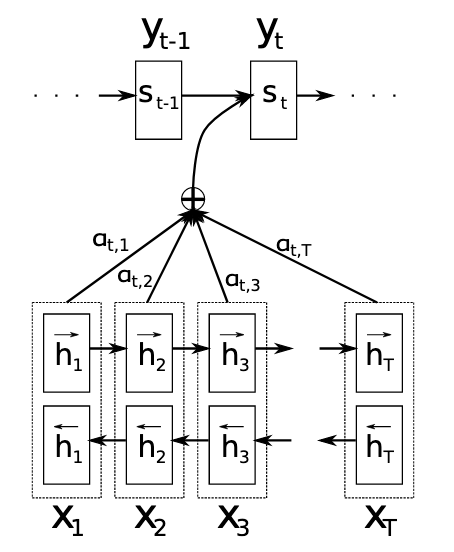}}
\end{center}
\caption {Additive attention mechanism from \cite{bahadanau}}
\vspace{0mm}
\label{bahadanau}
\end{figure}
    \[ where\ the\ decoder\ state, \ s_t = f(s_{t-1}, y_{t-1}, c_t)\]
    \[ and,\ \ h_i = [\vec{h_i}^{\,T}; \cev{h_i}^{\,T}]^{\,T}\]
\begin{equation} \label{eq:1}
    c_t = \sum_{i=1}^{n} \alpha_{t, i}h_i 
\end{equation}
\begin{equation} \label{eq:2}
    \alpha_{t, i} = align(y_t, x_i) 
\end{equation}
\begin{equation} \label{eq:3}
    align(y_t, x_i) = softmax(score(s_{t-1}, h_i)) 
\end{equation}
\begin{equation}  \label{eq:4}
    score(s_t, h_i) = v_a^T \tanh(w_a [ s_t; h_i ])
\end{equation}

\cite{bahadanau} states that attention model was trained on WMT’14 English to French dataset. Two models have been trained to compare the performance of the sequence to sequence architecture with (RNNsearch) and without (RNNencdec) attention. The RNNencdec has 1000 hidden units in encoder as well as decoder and RNNsearch has a bidirectional 1000 unit layer in the encoder and 1000 hidden unit layer at decoder. RNNsearch outperforms RNNencdec with BLEU\cite{bleu} score of 34.16 compared to the RNNencdec with a score of 26.71 with 50 as sentence length and no unknown ('UNK') words. Other results include RNNsearch: BLEU score- 31.44 and RNNencdec: BLEU score- 24.19 with sentence length 30\cite{bahadanau}.

\subsection{Self Attention}
Bahadanau Attention\cite{bahadanau} leads to high computational complexity due to its sequential nature. This was marked by computing a separate context vector for every state of the decoder. Also, from equation (3) it is observed that due to the sequential nature of RNNs, i.e. h\textsubscript{2} will be computed after h\textsubscript{1}, there can't be any parallelism in computation. These problems are addressed by self-attention. As stated in \cite{transformer}, self-attention has been used in various use cases viz reading comprehension, abstractive text summarization, textual entailment and learning task-independent sentence representations\cite{self1}, \cite{self2}, \cite{self3}. The transformer \cite{transformer} too uses self attention which is discussed in the next section.

\begin{figure}[!h]
\begin{center}
\scalebox{0.18}{\includegraphics{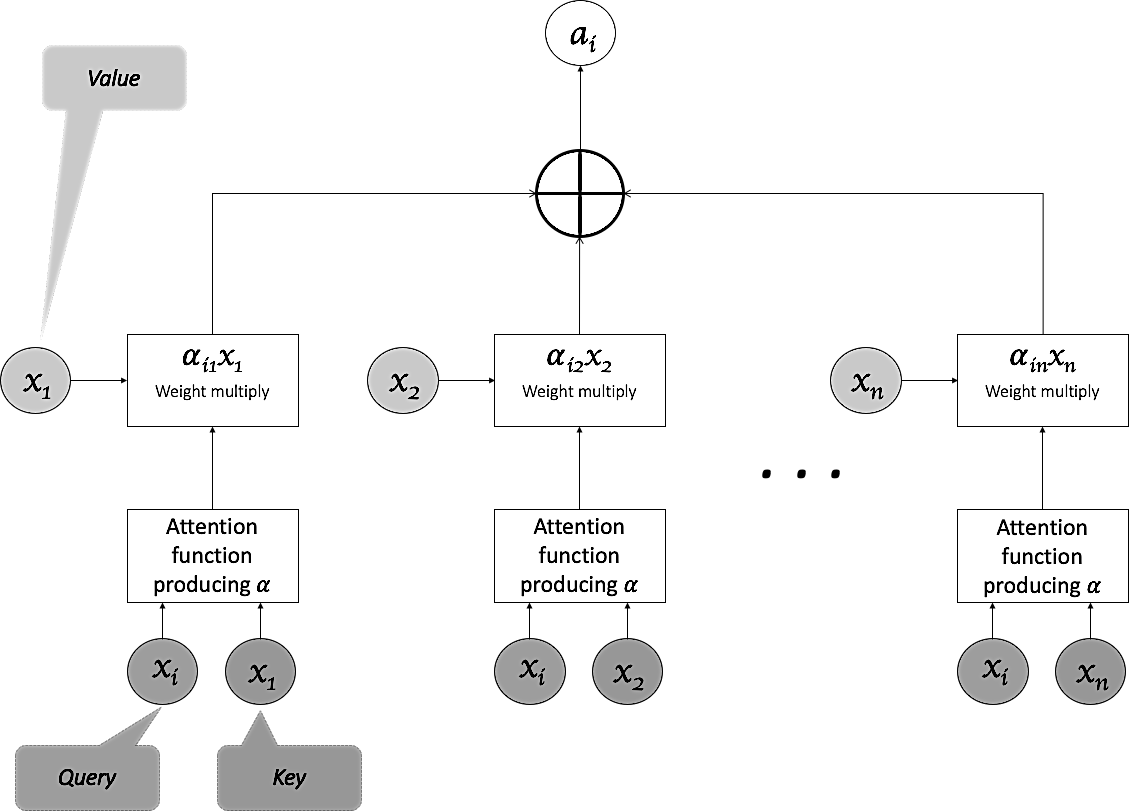}}
\caption {Self Attention}
\vspace{0mm}
\end{center}
\label{self_attn1}
\end{figure}

The attention mechanism can be visualised as an operation of a key (input vectors), query (last decoder hidden state), value (output of the layer) wherein the query and the key give the weights which are used to take a weighted sum with values to give the final state vector. Self attention aims to replace the sequential structure of RNNs (depending on previous state for the next state) by replacing RNNs. Instead of using hidden states for the inputs, in self attention, directly the inputs are used. As shown in figure 4, the key and value are x\textsubscript{1}, x\textsubscript{2}, ... and they are operated with the query x\textsubscript{i} which belong to the same sequence; hence "Self" Attention.

\cite{transformer} points out at \cite{self1}, \cite{self2}, \cite{self3}, \cite{self4} which are some of the papers that have used self attention. In \cite{self1}, to elicit a sense of memory in regular LSTM, an internal neural network is used to preserve the memory from a few previous states as well. This is called a LSTMN\cite{self1}. For sentiment analysis, Stanford Sentiment Treebank dataset was used where LSTMN got 87\% accuracy in binary sentiment analysis. Further, on experiments like Natural Language Inference, the dataset is Stanford Natural Language Inference (SNLI) dataset, where the accuracy achieved was 86.3\% for test data. \cite{self2} is specific to Natural Language Inference and it uses a Attend-Compare-Aggregate order the mechanism is called intra-sentence attention. The accuracy on SNLI dataset is 86.8\%. \cite{self3} focuses on sentence embedding using LSTMs and self attention. Results obtained on 3 different datasets for 3 separate tasks viz Twitter tweets for author age prediction gives 80.45\% accuracy, Yelp for sentiment analysis gives 64.21\% accuracy and SNLI for Textual entailment yields 84.4\% which is near state of the art. \cite{self4} essentially combines the pros of machine learning techniques and reinforcement techniques incorporated in the encoder-decoder architecture\cite{seq2seq} with intra-attention (self attention) to improve the performance of text summarization. The results are obtained for CNN/Daily Mail and New York Times datasets. The metric used is called ROGUE score\cite{rouge} which is discussed in later sections. Obtained results for ML+RL+intra attention 42.94 R-1 score and 26.02 R-2 score which beats the state of the art. Another special mention of the Self attention mechanism is \cite{sa_gan} which uses GANS\cite{gan} with self attention to achieve outstanding performance in image generation, however the emphasis on this is out of scope for this discussion.

\subsection{Multi-step Attention}
\cite{convseq} has implemented a block structure for extracting text features in a sequential manner where each block consists of a 1-D Convolution layer. To enable Deep Convolutional networks, \cite{convseq} includes residual / skip connections\cite{resnet}. For decoder, a separate attention mechanism is introduced. The current encoder state $h\textsubscript{i}\textsuperscript{l}$ is combined with target element $g\textsubscript{i}$ for achieving attention. The mathematical representation for multi-step attention from \cite{convseq} is:

\begin{equation} \label{eq:5}
    {d_i}^{\,l} = {W_d}^{\, l} {h_i}^{\, l} + {b_d}^{\,l} + {g_i}
\end{equation}
\begin{equation} \label{eq:6}
    {a}_{ij}^{\, l} = softmax({d_i}^{\,l}.{z_t}^{\, u})
\end{equation}
\begin{equation} \label{eq:7}
    {c_i}^{\, l} = \sum_{j=1}^{m}{a}_{ij}^{l} ({z_j}^u + {e_j})
\end{equation}
\begin{figure}[!h]
\begin{center}
\scalebox{0.45}{\includegraphics{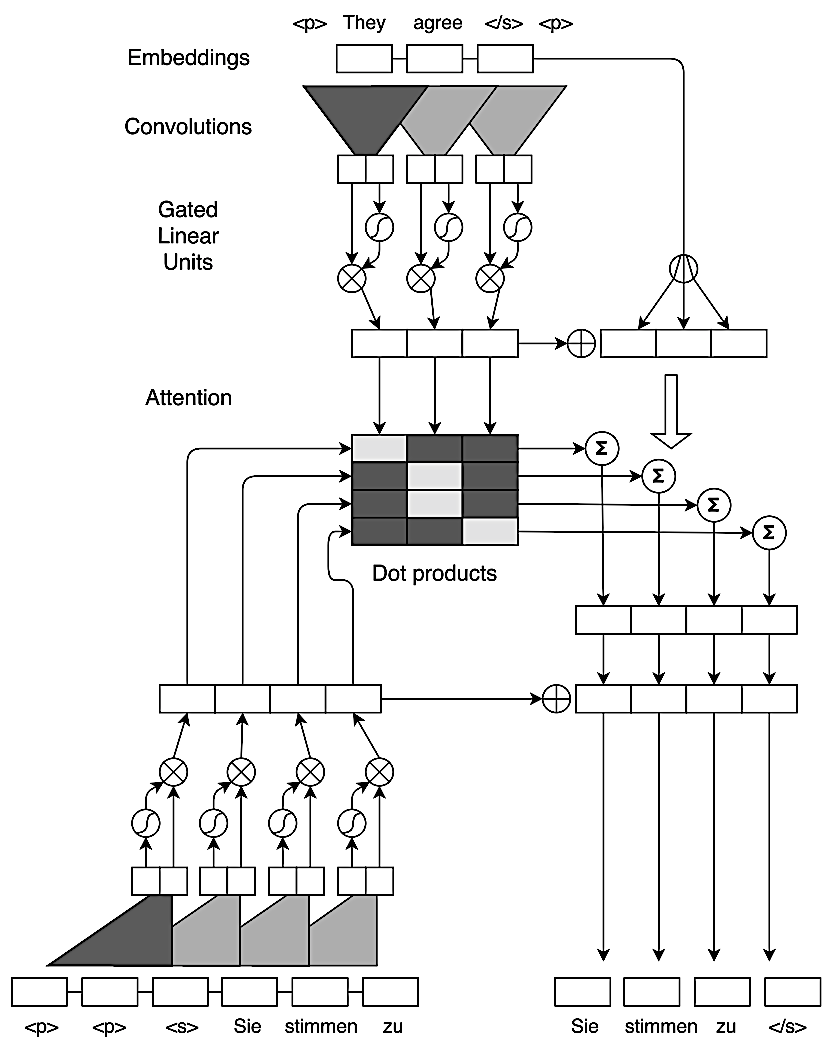}}
\caption {Convolutional Sequence to Sequence Learning Architecture from \cite{convseq}}
\vspace{0mm}
\end{center}
\label{self_attn2}
\end{figure}

With respect to NMT, the architecture is experimented on several datasets including WMT’16 English-Romanian (sequence length: 175, 200k vocabulary size), WMT’14 English-German (same as \cite{effective_attn}), WMT’14 English-French (sequence length: 175). The results obtained include BLEU 30.02 on WMT’16 English-Romanian, BLEU: 25.16 on WMT’14 English-German, BLEU 40.51 on WMT’14 English-French.

\section{Pointer Softmax}
In NMT many of the words in the corpus are generalised by the '<unk>' (unknown) representation so as to reduce the vocabulary size to scale and discard very less frequent words from the vocabulary. However in many cases this depletes the performance of the model. In \cite{pointer_softmax}, the architecture used is plain encoder-decoder\cite{seq2seq} with attention\cite{bahadanau} and there are two softmax layers to predict the next word in the conditional language model; one to predict the location of the word in the model and the second to predict the word in the shortlist vocabulary. The decision of using which softmax is made by an Multi-Layered Perceptron or Feed Forward Neural Net. The intuition to this nmechanism is the human tendency to pointing at objects when not knowing their name. 

\begin{figure}[!h]
\begin{center}
\scalebox{0.5}{\includegraphics{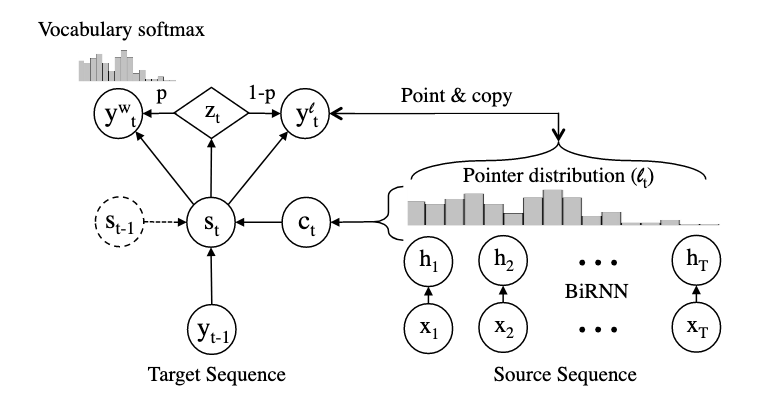}}
\caption {Pointer Softmax\cite{pointer_softmax} with Encoder-Decoder\cite{seq2seq} + Attention\cite{bahadanau}}
\vspace{0mm}
\end{center}
\label{self_attn3}
\end{figure}

The model was tested on newstest2011 French to English dataset and gives better results (BLEU score 23.76) as compared to traditional attention based NMT model (20.19) on the same dataset for sequence length 50. 30,000 tokens for both the source and target language shortlist vocabularies were used. The corpus contains 134,831 English and 153,083 French  words. A word level translation dictionary is maintained for 15,953 words that belong to none of the vocabularies. About 49,490 words are shared between English and French corpora of Europarl.

\section{Google's Neural Machine Translation System}
GNMT\cite{gnmt} introduces a parallelism in model training and speeding up the training process. Along with parallelism, GNMT also introduces noble segmentation approaches to address out of vocabulary (OOV) words including wordpiece model which was originally developed o solve a Japanese/Korean segmentation problem for the Google speech recognition system\cite{segmentation}; and the mixed word/character model.

\begin{figure}[!h]
\begin{center}
\scalebox{0.26}{\includegraphics{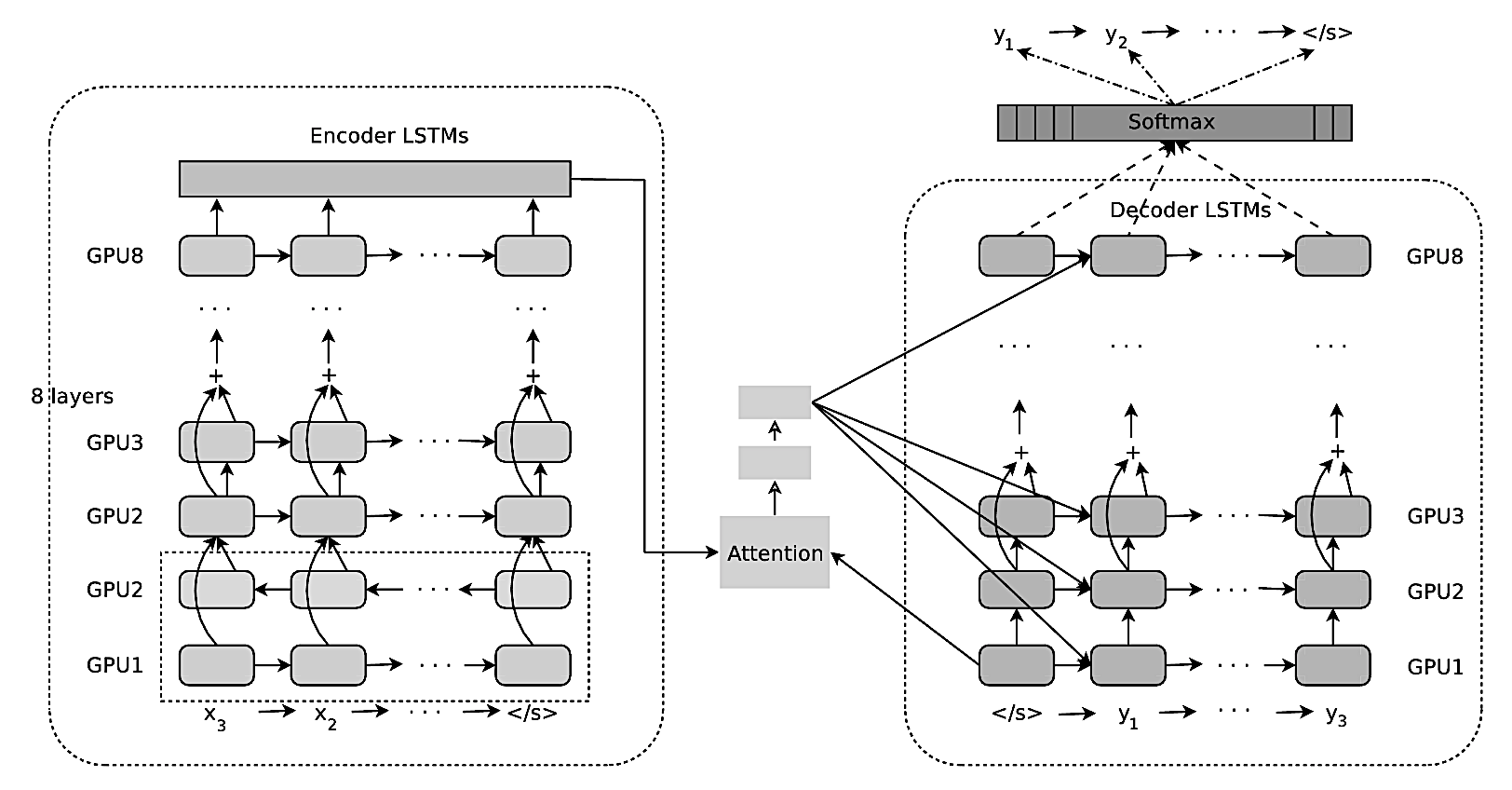}}
\caption {GNMT Architecture \cite{gnmt}}
\vspace{0mm}
\end{center}
\label{gnmt}
\end{figure}

The first layer of encoder is a bi-directional layer, which computes on a GPU in parallel first and then the rest of the 7 unidirectional layers are trained in parallel on separate GPUs. The output is fed to the attention module which passes the output to the 8 decoder layers. The model consists of residual connections so as to enable training very deep neural network architectures\cite{resnet}. The first encoder layer is bidirectional as some of the grammars in different languages exhibit positional resemblance in the words and hence feeding the same input in reverse sometime helps.
\par
GNMT was trained on the publicly available WMT'14 English-German as well as English-French datasets. The results are compared to several models as well as human evaluation. On the En-Fr with WPM-32K, the BLEU score was 38.95 and with Mixed Word/Character model it was 38.39. On En-De with WPM-32K the BLEU was 24.61 and with Mixed Word/Character model it was 24.17. For model ensemble, WPM-32K (8 models) on En-Fr had BLEU 40.35 and for RL-refined model, BLEU was 41.16; for En-De, the WPM-32K BLEU with 8 models was 26.20 and RL-refined WPM-32K (8 models) was 26.30. Human side-by-side evaluation scores of WMT En-Fr models has an average score of 4.82 whereas the NMT model with and without RL has an average score of 4.44 and 4.46 respectively.

\section{The Transformer}
\cite{transformer} proposes the well known transformer model. This is one of the most successful implementations of Self-Attention. Although, the transformer introduces a scaled-dot product mechanism paired with multi-headed self-attention. The architecture proposed in \cite{transformer} is:

\begin{figure}[!h]
\begin{center}
\scalebox{0.3}{\includegraphics{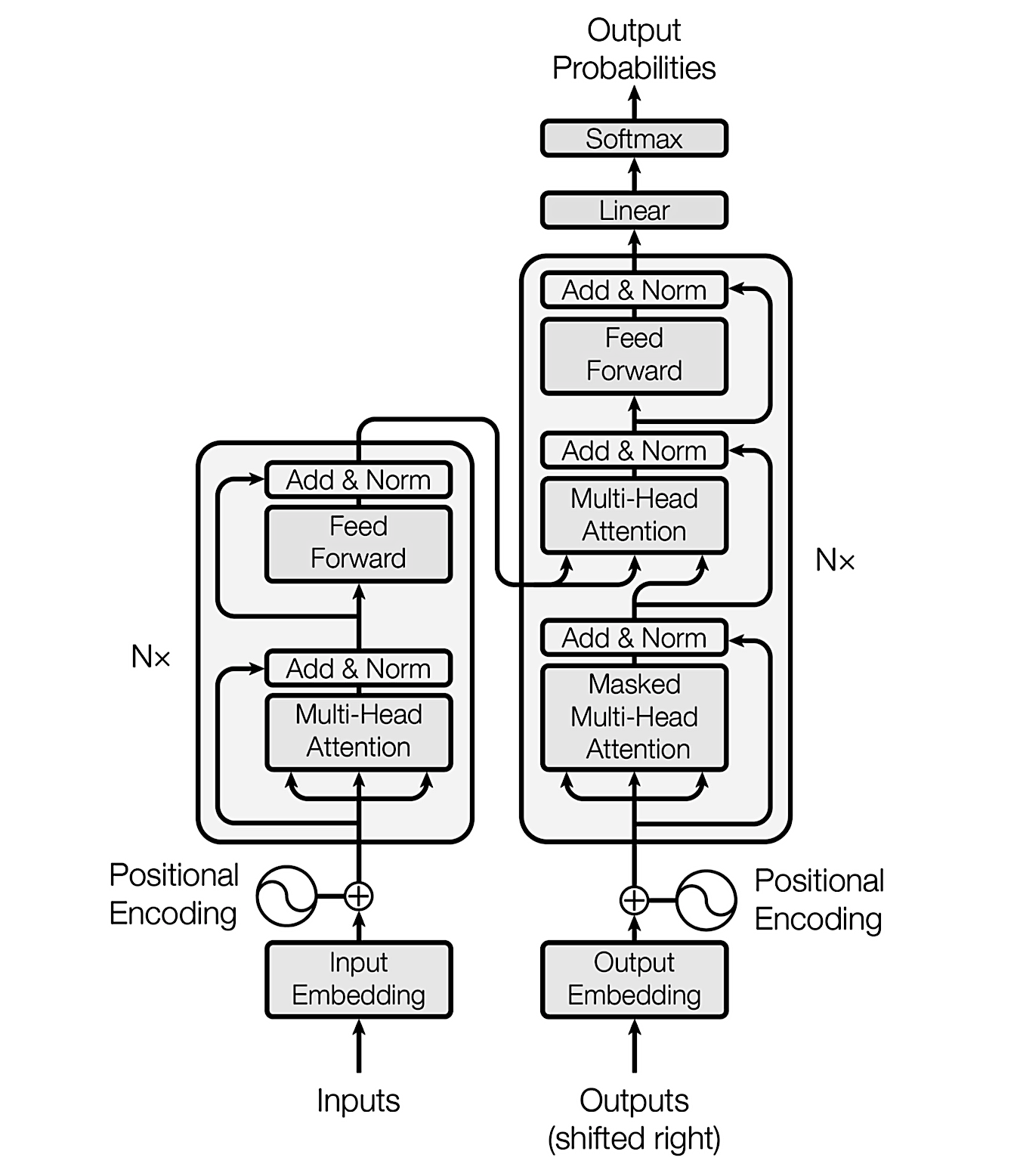}}
\caption {Transformer from Attention is all you need \cite{transformer}}
\vspace{0mm}
\end{center}
\label{transformer1}
\end{figure}

The major modules of the transformer model that make up the architecture include the encoder and decoder multi-head attention, feed forward respectively. The inputs initially are the word embeddings of the words being processed by the model and the inputs to the subsequent layers are the outputs of the preceding layers. The pictorial representation of scaled dot-product and multi-head attention is as follows:
\begin{figure}[!h]
\begin{center}
\scalebox{0.2}{\includegraphics{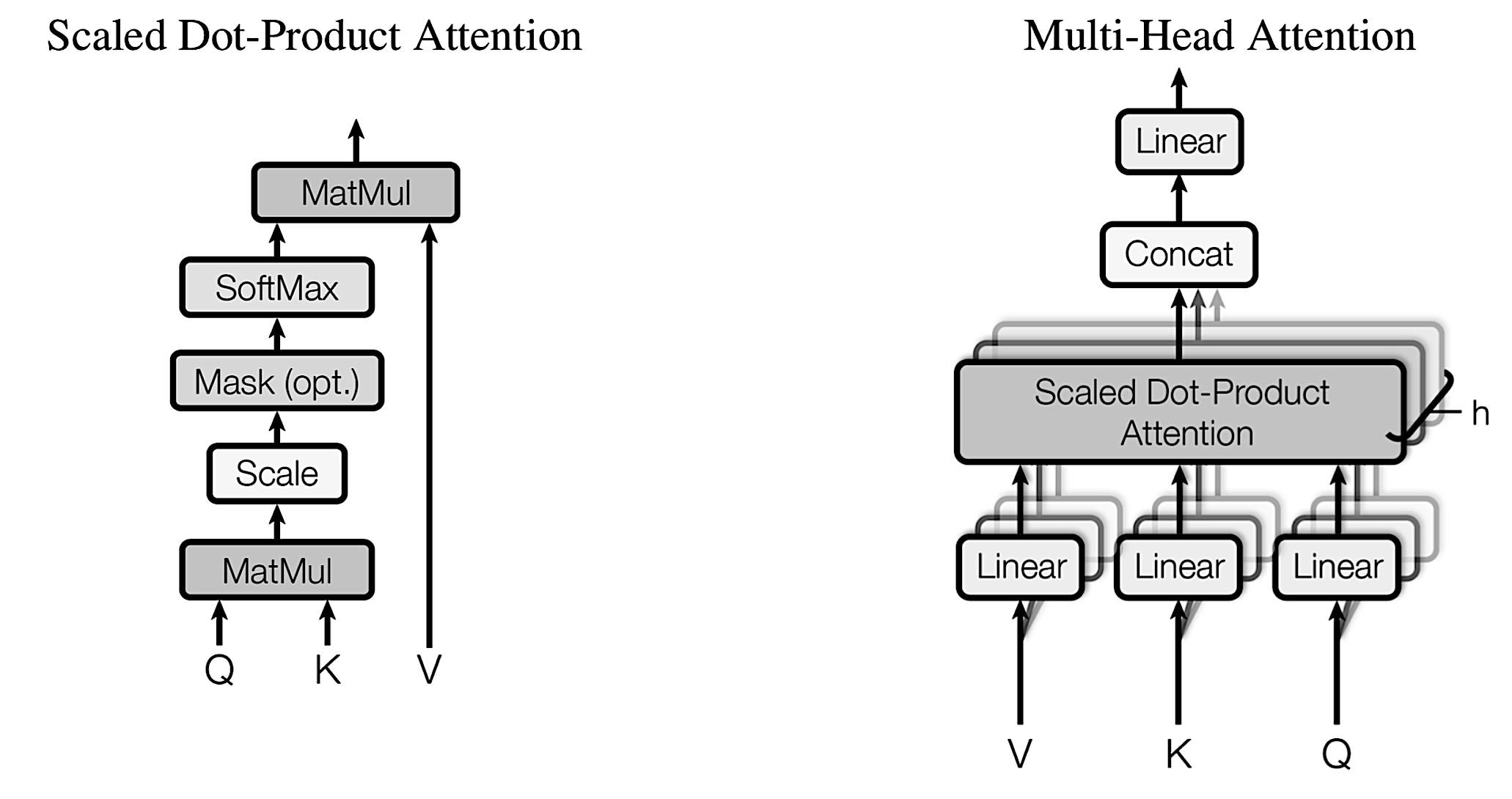}}
\caption{Scaled Dot-Product and Multi-Head Attention from\cite{transformer}}
\vspace{0mm}
\end{center}
\label{transformer2}
\end{figure}

One head of the attention module includes self attention with a scaling add-on; the Key-Query-Value structure stays. The Query and Key dot product is scaled by a value $\sqrt{d_k}$ where $d$ is the dimension of the input and is merely a scaling parameter to avoid explosion of values after the dot-product. The dot-product is then softmax-ed and operated with the value vector. This is essentially the math for calculation of the vector ${a_i}$ from equation (2) in the self-attention fashion. However, in transformer multiple ${a_i}$s are calculated with different weights. This gives a notion of extracting multiple feature maps as in Convolutional Neural Networks\cite{cnn}. These vectors are then concatenated and then fed to a feed forward neural network layer. Similarly at the decoder the expected output is fed to the decoder layer and then combined at the 'multi-head-encoder-decoder-attention' layer with the encoder outputs.
\par
The transformer was trained on WMT'14 English-German dataset, 4.5 million sentence pairs. Sentences were encoded using byte-pair encoding \cite{nmt_architectures}, which has a vocabulary of about 37000 tokens at source and targets. For English-French, WMT'14 English-French dataset with of 36M sentences was used. 32000 word-piece vocabulary was used \cite{gnmt}. The transformers yielded state of the art results with the big transformer model scoring 28.4 BLEU on English-German dataset and 41 BLEU on English-French dataset. The transformer model aims not only at achieving state of the art results but also to do that with minimal computational complexity. To measure performance, \cite{transformer} has calculated FLOPs (Floating Point Operations Per Second) which is 10-100 times better than the state of the art models.

\section{BERT in NMT}
Bidirectional Encoder Representations from Transformers\cite{bert1} is google's state of the art language modelling architecture based on the transformers model\cite{transformer}. BERT essentially encourages the recent trends of transfer learning. The architecture is trained on bidirectional (apparently non-directional) transformer which tends to remove the benchmarks set by the recurrent models. Since the goal was to train a language model that could be modularized into the requirement by fine tuning the pretrained model, BERT essentially is a language model that introduces a novel approach called MLM (Masked LM). This involves 'masking' 15\% random words in a given sentence and using the bidirectional model to predict them.

\begin{figure}[!h]
\begin{center}
\scalebox{0.22}{\includegraphics{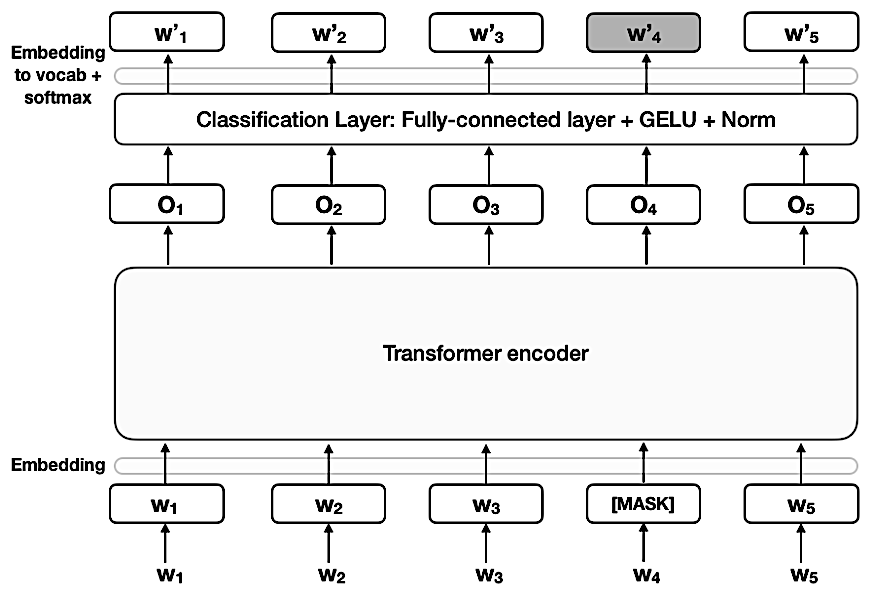}}
\caption{BERT \cite{bert1}}
\vspace{0mm}
\end{center}
\label{bert1}
\end{figure}

While training Next Sentence Prediction (NSP) in BERT, two inputs are fed to the model, the source sentence (current sentence) and the target sentence (next sentence). For this, the authors of \cite{bert1} trained with 50\% of the sentences trained with wrong target sentences (i.e. a sentence which is out of context). The reason behind this being that the model should learn to be incisive in telling if the sentence belongs in the context of the given context.
\par
BERT fine tuning implies appending an use case on the pretrained language model. \cite{bert2}, \cite{bert3} discusses implementation of BERT at NMT. In  \cite{bert2} a concerted training framework(CTNMT) is introduced that serves the integration of the pretrained language model with NMT. CTMT consists of 3 modules viz. asymptotic distillation to make sure that the NMT model retains the previous pretrained context; dynamic switching gate to avoid forgetting of pretrained context; strategy to adjust the learning paces according to a scheduled policy\cite{bert2}. In asymptotic distillation, \cite{bert2} essentially the mean squared error between the final hidden state of the language model and the NMT model.

\begin{equation} \label{eq:8}
    L_{kd} = || \hat{h}^{lm} - h_l ||_{2}^{2}
\end{equation}

This loss is used in conjunction with the traditional cross-entropy loss of the NMT model by a weighted average on a hyper-parameter $\alpha$. This can be summarized mathematically as:

\begin{equation} \label{eq:9}
    L = \alpha L_{nmt} + (1 - \alpha) L_{kd}
\end{equation}

To address the problem of forgetting either the language model context or the NMT context, \cite{bert2} introduces dynamic switch, which involves taking a weighted average of the hidden states of the language model and NMT respectively. The mathematical representation of the final hidden state is:

\begin{equation} \label{eq:10}
    g = \sigma(Wh^{lm} + Uh^{nmt} + b)
\end{equation}
\begin{equation} \label{eq:11}
    h = g \cdot h^{lm} + (1 - g) \cdot h^{nmt}
\end{equation}

In addition to the aforementioned methods, the paper also introduces rate scheduled learning rate strategy to tune each component with different learning rate.

\begin{equation} \label{eq:12}
    \theta^{lm}_{t} = \theta^{lm}_{t - 1} - \eta^{lm}\nabla_{\theta^{lm}}L(\theta^{lm})
\end{equation}
\begin{equation} \label{eq:13}
    \theta^{nmt}_{t} = \theta^{nmt}_{t - 1} - \eta^{nmt}\nabla_{\theta^{nmt}}L(\theta^{nmt})
\end{equation}

The key is to first converge the NMT model and then jointly train it with the language model and moderately tune only the NMT parameters to make sure the language model data isn't forgotten.
\par
The dataset to evaluate CTNMT used is WMT English-German translation task and also English-French, English-Chinese. CTNMT with only rate scheduling obtained a BLEU score of 29.7 on En-De dataset which outperforms the transformer (big) with almost 1 BLEU point. The same model with En-Fr and En-Zh yielded BLEU of 41.6 and 38.4 respectively. WIth CTMT and dynamic switch En-De has a BLEU score of 29.4, with En-Fr 41.6 and 38.6 with En-Zh. With Asymptotic Distillation, CTNMT gives 29.2, 41.6 and 38.3 on En-De, En-Fr, En-Zh respectively. Finally, with all the methodologies combined, BERT CTNMT gives state of the art results with BLEU scores 30.1, 42.3 and 38.9 on En-De, En-Fr and En-Zh respectively.

\section{Evaluation Metric: BLEU}
BLEU \cite{bleu} is the most widely used evaluation metric in NMT tasks. It is marked by simplicity in calculation and closeness to human calculation. The cardinal task in calculating a BLEU score is simply comparing the reference sequence n-gram with the candidate sequence n-gram. BLEU shows resemblance to calculating precision score. To calculate precision, one may take the number of tokens in the candidate sequence that occur in the reference sequence and divide it to total number of tokens in the candidate sequence. However, candidates can overgenerate reasonable words and hence produce a high precision score. Thus, arises the need for exhausting a given reasonable token once it is matched to a token in the reference sequence. This is explained in the modified unigram precision method in \cite{bleu}. Additionally, the proposed metric uses combined modified n-gram precision along with sentence brevity penalty. The calculation of BLEU can be summarized as :

First the mean of the test corpus' modified precision scores is taken. Then the brevity penalty is calculated.
\begin{equation} \label{eq:14}
    BP =    \begin{cases*}
            1 & $if\ c > r$ \\
            e ^ {(1 - r/c)} & $if\ c \leq r$ \\
            \end{cases*}
\end{equation}

\begin{equation} \label{eq:15}
    BLEU = BP \cdot \exp{\bigg(\sum_{n=1}^{N}w_n \log{p_n}\bigg)}
\end{equation}

In addition to BLEU there are several metrics mentioned in this discussion viz. ROUGE score \cite{rouge} to evaluate summarizers and glue score \cite{glue} to evaluate language models. Although these are out of scope for this discussion.

\section{Results}
In this section, results from various previously discussed models are compared. In most of the approaches, the dataset used for evaluation is the WMT'14 En-Fr and En-De. other datasets used are WMT'16 English-Romanian, newtest2011 Fr-En, WMT'14 En-Zh. Along with these, some models are discussed that address to other use cases to delineate the Self-Attention Mechanism; these include evaluation on Stanford Treebank Dataset for binary sentiment analysis, Stanford Natural Language Inference (SNLI) Dataset for NLI \cite{self1}, \cite{self2} and Textual Entailment\cite{self3}, Twitter Tweets for author age prediction\cite{self3}, Yelp for sentiment analysis\cite{self3} and CNN Daily Mail and New York Times data for abstractive text summarization\cite{self4}. The results obtained in the models discussed so far are summarized in Table I.

\begin{table*}
    \centering
    \caption{Results Comparisons of Major NMT Models (Evaluation Metric: BLEU)}
    \label{tab:1}
    \renewcommand{\arraystretch}{1.5}
    \begin{tabular}{|c|c|c|c|c|c|}
        \hline
        \textbf{Model} & \textbf{En-Fr} & \textbf{En-De} & \textbf{En-Zh} & \textbf{En-Rom} & \textbf{Fr-En}\\
        \hline
        \textbf{seq2seq (5 deep LSTMs, 384M parameters, 8k dimensions)} & 34.81 & - & - & - & - \\
        \hline
        \textbf{bahadanau attention (1000 units in enc and dec)} & 34.16 & - & - & - & - \\
        \hline
        \textbf{Convolutional seq2seq (seq length 175, vocab 200k)} & 40.51 & 25.16 & - & \textbf{30.02} & - \\
        \hline
        \textbf{Pointer Softmax (seq length 50, 30k tokens at source and target)} & - & - & - & - & \textbf{23.76} \\
        \hline
        \textbf{GNMT (WPM 32k)} & 38.95 & 24.61 & - & - & - \\
        \hline
        \textbf{GNMT (Mixed Word/Character Model)} & 38.39 & 24.17 & - & - & - \\
        \hline
        \textbf{GNMT (WPM 32k ensemble of 8 models)} & 40.35 & 26.20 & - & - & - \\
        \hline
        \textbf{GNMT (RL refined ensemble model)} & 41.16 & 26.30 & - & - & - \\
        \hline
        \textbf{GNMT (Human side by side evaluated with RL average score)} & 4.44 & - & - & - & - \\
        \hline
        \textbf{GNMT (Human side by side evaluated without RL average score)} & 4.46 & - & - & - & - \\
        \hline
        \textbf{GNMT (Human evaluation average score)}
        \textbf{Transformer (base model)} & 38.1 & 27.3 & - & - & - \\
        \hline
        \textbf{Transformer (big)} & 41.0 & 28.4 & - & - & - \\
        \hline
        \textbf{BERT (CTNMT + Transformer (base))} & 41.0 & 27.2 & 37.3 & - & - \\
        \hline
        \textbf{BERT (CTNMT + Rate Scheduling)} & 41.6 & 29.7 & 38.4 & - & - \\
        \hline
        \textbf{BERT (CTNMT + Dynamic Switch)} & 41.4 & 29.4 & 38.6 & - & - \\
        \hline
        \textbf{BERT (CTNMT + Asymptotic Distillation)} & 41.6 & 29.2 & 38.3 & - & - \\
        \hline
        \textbf{BERT (CTNMT + All)} & \textbf{42.3} & \textbf{30.1} & \textbf{38.9} & - & - \\
        \hline
        
    \end{tabular}
\end{table*}

\section{Related Work}
The comparisons in Table I clearly depict the evolution in Neural Machine Translation. However, certain additions to the current state of the art models may apparently increase the performance by decent figures. One add-on can be integrating the Memory Networks architecture which has been discussed in \cite{memnet1}, \cite{memnet2}, \cite{memnet3}. The intuition to this can be obtained by visualising as an external data structure implemented to maintain the contextual information and using a weighted sum approach like in any other attention mechanism select the most relevant parts from those context data structures. For example, in case of \cite{bert2}, an external data structure can keep the language model context for various similar sequences and a weighted average / sum can be taken from these external sources followed by application of the approaches implemented for forgetting NMT / LM contexts.

\section{CONCLUSION}
With Neural Machine Translation, almost all other (statistical, probabilistic, etc.) models for Machine Translation have been ruled out. To summarize, the approaches started with RNN-based methodologies which was later replaced by non-RNN methodologies which included considering all the tokens in a given sequence to preserve long term context and in a way eliciting an end-to-end notion in the model architecture. The state of the art language model, BERT has outperformed almost all the existing models in most of the NLP tasks by fine tuning the pretrained language model. The CTNMT model from \cite{bert2} has outperformed all the discussed models in NMT tasks by appreciable differences in the evaluation scores. Neural Machine Translation tasks have use cases in a wide range of use cases including Government, Education, Medical domains and these demand exacting results. These models certainly can meet the expectations with room for improvement of course.

\bibliography{references}

\begin{thebibliography}{10}

\bibitem{vanishing_grad}
R.~{Pascanu}, T.~{Mikolov}, and Y.~{Bengio}, ``{On the difficulty of training
  Recurrent Neural Networks},'' {\em arXiv e-prints}, p.~arXiv:1211.5063, Nov.
  2012.

\bibitem{seq2seq}
I.~{Sutskever}, O.~{Vinyals}, and Q.~V. {Le}, ``{Sequence to Sequence Learning
  with Neural Networks},'' {\em arXiv e-prints}, p.~arXiv:1409.3215, Sept.
  2014.

\bibitem{bleu}
K.~Papineni, S.~Roukos, T.~Ward, and W.-J. Zhu, ``Bleu: A method for automatic
  evaluation of machine translation,'' in {\em Proceedings of the 40th Annual
  Meeting on Association for Computational Linguistics}, ACL ’02, (USA),
  p.~311–318, Association for Computational Linguistics, 2002.

\bibitem{bahadanau}
D.~{Bahdanau}, K.~{Cho}, and Y.~{Bengio}, ``{Neural Machine Translation by
  Jointly Learning to Align and Translate},'' {\em arXiv e-prints},
  p.~arXiv:1409.0473, Sept. 2014.

\bibitem{transformer}
A.~{Vaswani}, N.~{Shazeer}, N.~{Parmar}, J.~{Uszkoreit}, L.~{Jones}, A.~N.
  {Gomez}, L.~{Kaiser}, and I.~{Polosukhin}, ``{Attention Is All You Need},''
  {\em arXiv e-prints}, p.~arXiv:1706.03762, June 2017.

\bibitem{self1}
J.~{Cheng}, L.~{Dong}, and M.~{Lapata}, ``{Long Short-Term Memory-Networks for
  Machine Reading},'' {\em arXiv e-prints}, p.~arXiv:1601.06733, Jan. 2016.

\bibitem{self2}
A.~P. {Parikh}, O.~{T{\"a}ckstr{\"o}m}, D.~{Das}, and J.~{Uszkoreit}, ``{A
  Decomposable Attention Model for Natural Language Inference},'' {\em arXiv
  e-prints}, p.~arXiv:1606.01933, June 2016.

\bibitem{self3}
Z.~{Lin}, M.~{Feng}, C.~{Nogueira dos Santos}, M.~{Yu}, B.~{Xiang}, B.~{Zhou},
  and Y.~{Bengio}, ``{A Structured Self-attentive Sentence Embedding},'' {\em
  arXiv e-prints}, p.~arXiv:1703.03130, Mar. 2017.

\bibitem{self4}
R.~{Paulus}, C.~{Xiong}, and R.~{Socher}, ``{A Deep Reinforced Model for
  Abstractive Summarization},'' {\em arXiv e-prints}, p.~arXiv:1705.04304, May
  2017.

\bibitem{rouge}
C.-Y. Lin, ``{ROUGE}: A package for automatic evaluation of summaries,'' in
  {\em Text Summarization Branches Out}, (Barcelona, Spain), pp.~74--81,
  Association for Computational Linguistics, July 2004.

\bibitem{sa_gan}
H.~{Zhang}, I.~{Goodfellow}, D.~{Metaxas}, and A.~{Odena}, ``{Self-Attention
  Generative Adversarial Networks},'' {\em arXiv e-prints},
  p.~arXiv:1805.08318, May 2018.

\bibitem{gan}
I.~J. {Goodfellow}, J.~{Pouget-Abadie}, M.~{Mirza}, B.~{Xu}, D.~{Warde-Farley},
  S.~{Ozair}, A.~{Courville}, and Y.~{Bengio}, ``{Generative Adversarial
  Networks},'' {\em arXiv e-prints}, p.~arXiv:1406.2661, June 2014.

\bibitem{convseq}
J.~{Gehring}, M.~{Auli}, D.~{Grangier}, D.~{Yarats}, and Y.~N. {Dauphin},
  ``{Convolutional Sequence to Sequence Learning},'' {\em arXiv e-prints},
  p.~arXiv:1705.03122, May 2017.

\bibitem{resnet}
K.~{He}, X.~{Zhang}, S.~{Ren}, and J.~{Sun}, ``{Deep Residual Learning for
  Image Recognition},'' {\em arXiv e-prints}, p.~arXiv:1512.03385, Dec. 2015.

\bibitem{effective_attn}
M.-T. {Luong}, H.~{Pham}, and C.~D. {Manning}, ``{Effective Approaches to
  Attention-based Neural Machine Translation},'' {\em arXiv e-prints},
  p.~arXiv:1508.04025, Aug. 2015.

\bibitem{pointer_softmax}
C.~{Gulcehre}, S.~{Ahn}, R.~{Nallapati}, B.~{Zhou}, and Y.~{Bengio},
  ``{Pointing the Unknown Words},'' {\em arXiv e-prints}, p.~arXiv:1603.08148,
  Mar. 2016.

\bibitem{gnmt}
Y.~{Wu}, M.~{Schuster}, Z.~{Chen}, Q.~V. {Le}, M.~{Norouzi}, W.~{Macherey},
  M.~{Krikun}, Y.~{Cao}, Q.~{Gao}, K.~{Macherey}, J.~{Klingner}, A.~{Shah},
  M.~{Johnson}, X.~{Liu}, {\L}.~{Kaiser}, S.~{Gouws}, Y.~{Kato}, T.~{Kudo},
  H.~{Kazawa}, K.~{Stevens}, G.~{Kurian}, N.~{Patil}, W.~{Wang}, C.~{Young},
  J.~{Smith}, J.~{Riesa}, A.~{Rudnick}, O.~{Vinyals}, G.~{Corrado},
  M.~{Hughes}, and J.~{Dean}, ``{Google's Neural Machine Translation System:
  Bridging the Gap between Human and Machine Translation},'' {\em arXiv
  e-prints}, p.~arXiv:1609.08144, Sept. 2016.

\bibitem{segmentation}
M.~Schuster and K.~Nakajima, ``Japanese and korean voice search,'' in {\em
  International Conference on Acoustics, Speech and Signal Processing},
  pp.~5149--5152, 2012.

\bibitem{cnn}
Y.~Lecun, L.~Bottou, Y.~Bengio, and P.~Haffner, ``Gradient-based learning
  applied to document recognition,'' {\em Proceedings of the IEEE}, vol.~86,
  pp.~2278 -- 2324, 12 1998.

\bibitem{nmt_architectures}
D.~Britz, A.~Goldie, M.-T. Luong, and Q.~Le, ``Massive exploration of neural
  machine translation architectures,'' in {\em Proceedings of the 2017
  Conference on Empirical Methods in Natural Language Processing}, (Copenhagen,
  Denmark), pp.~1442--1451, Association for Computational Linguistics, Sept.
  2017.

\bibitem{bert1}
J.~{Devlin}, M.-W. {Chang}, K.~{Lee}, and K.~{Toutanova}, ``{BERT: Pre-training
  of Deep Bidirectional Transformers for Language Understanding},'' {\em arXiv
  e-prints}, p.~arXiv:1810.04805, Oct. 2018.

\bibitem{bert2}
J.~Yang, M.~Wang, H.~Zhou, C.~Zhao, Y.~Yu, W.~Zhang, and L.~Li, ``Towards
  making the most of bert in neural machine translation,'' {\em ArXiv},
  vol.~abs/1908.05672, 2019.

\bibitem{bert3}
S.~{Clinchant}, K.~W. {Jung}, and V.~{Nikoulina}, ``{On the use of BERT for
  Neural Machine Translation},'' {\em arXiv e-prints}, p.~arXiv:1909.12744,
  Sept. 2019.

\bibitem{glue}
A.~{Wang}, A.~{Singh}, J.~{Michael}, F.~{Hill}, O.~{Levy}, and S.~R. {Bowman},
  ``{GLUE: A Multi-Task Benchmark and Analysis Platform for Natural Language
  Understanding},'' {\em arXiv e-prints}, p.~arXiv:1804.07461, Apr. 2018.

\bibitem{memnet1}
J.~{Weston}, S.~{Chopra}, and A.~{Bordes}, ``{Memory Networks},'' {\em arXiv
  e-prints}, p.~arXiv:1410.3916, Oct. 2014.

\bibitem{memnet2}
S.~{Sukhbaatar}, A.~{Szlam}, J.~{Weston}, and R.~{Fergus}, ``{End-To-End Memory
  Networks},'' {\em arXiv e-prints}, p.~arXiv:1503.08895, Mar. 2015.

\bibitem{memnet3}
C.~{Xiong}, S.~{Merity}, and R.~{Socher}, ``{Dynamic Memory Networks for Visual
  and Textual Question Answering},'' {\em arXiv e-prints}, p.~arXiv:1603.01417,
  Mar. 2016.

\end{thebibliography}
\bibliographystyle{ieeetr}

\end{document}